\documentclass{article}

\usepackage{arxiv}

\usepackage[utf8]{inputenc} 
\usepackage[T1]{fontenc}    
\usepackage{hyperref}       
\usepackage{url}            
\usepackage{booktabs}       
\usepackage{amsfonts}       
\usepackage{nicefrac}       
\usepackage{microtype}      
\usepackage{lipsum}
\usepackage{amsmath}
\usepackage{tensor}

\title{A Gradient Estimator for Time-Varying Electrical Networks with Non-Linear Dissipation}

\author{
  Jack D. Kendall \\
  Rain Neuromorphics\\
  747 Broadway\\
  Redwood City, CA 94063 \\
  \texttt{jack@rain-neuromorphics.com} \\
}

\newtheorem{theorem}{Theorem}

\begin{document}
\maketitle

\begin{abstract}
We propose a method for extending the technique of equilibrium propagation \cite{scellier2017equilibrium} for estimating gradients in fixed-point neural networks to the more general setting of directed, time-varying neural networks by modeling them as electrical circuits. We use electrical circuit theory to construct a Lagrangian capable of describing deep, directed neural networks modeled using nonlinear capacitors and inductors, linear resistors and sources, and a special class of nonlinear dissipative elements called fractional memristors \cite{machado2013fractional}. We then derive an estimator for the gradient of the physical parameters of the network, such as synapse conductances, with respect to an arbitrary loss function. This estimator is entirely local, in that it only depends on information locally available to each synapse. We conclude by suggesting methods for extending these results to networks of biologically plausible neurons, e.g. Hodgkin-Huxley neurons \cite{hodgkin1952quantitative}.
\end{abstract}


\section{Introduction}
The question of how biological brains perform learning has attracted the attention of neuroscience researchers for many years. While intricate synaptic plasticity mechanisms are known (Hebbian, STDP, BCM, etc.), these mechanisms are all local in their effect \cite{hebb2005organization, dan2006spike, markram1997regulation, bienenstock1982theory}. In other words, the strength of a synapse is only affected by the activity of the two neurons it connects. This constraint is primarily of a physical origin, reflecting the physical structure of biological neural networks. 

It is an open question as to how these local synaptic weight changes can be coordinated globally across a multi-layered biological neural network. In deep learning, the coordinated optimization of all the synaptic weights in the network (even ones far from the output layer) is achieved via the backpropagation algorithm. However, how such coordination can be achieved in biological neural networks is currently unknown.

This open problem is popularly referred to as the \textbf{credit assignment problem}. Here, credit assignment refers to determining how much an arbitrary synapse contributed to the total reward or cost of an action or inference, and is the global signal that guides learning. It is strongly believed that it is effective credit assignment that allows artificial neural networks trained by deep learning to develop sophisticated internal representations \cite{richards2019deep}.

\subsection{Deep Learning}
Deep neural networks, i.e. artificial neural networks with many-layered or hierarchical structures, were long thought impossible to train using standard first-order learning algorithms such as stochastic gradient descent \cite{hochreiter1998vanishing}. However, the discovery of algorithmic techniques such as batch normalization and momentum coupled with architectural modifications such as residual connections have proven that deep neural networks can be trained to achieve state-of-the-art performance over a wide range of tasks including image and voice recognition, language translation, reinforcement learning, unsupervised learning, natural language processing, and others \cite{he2016deep, deng2013new, mnih2013playing, goodfellow2014generative}.

It is widely believed that this is due to a previously unexpected interaction between stochastic gradient descent (SGD) and the hierarchical loss surfaces of deep neural networks \cite{nguyen2018loss, li2017visualizing}. In short, the loss surface of a deep neural network can be shown to possess certain geometric properties which make SGD unlikely to be trapped in poor, locally optimal solutions. In fact, nearly all local optima found by SGD are very close in task performance to the globally optimal solution(s), and moreover tend to generalize better to unseen data than global optima \cite{zhu2018anisotropic}. This synergistic effect between stochastic gradient descent and the loss surfaces of modern deep neural networks not only explains their performance, but also suggests that deep (hierarchical) biological neural networks may also be using stochastic gradient descent to update their synaptic weights \cite{richards2019deep}.

The common rebuttal to this argument is that backpropagation, the algorithm used by deep learning practitioners to calculate the synapse gradients for use in SGD, is not biologically plausible \cite{stork1989backpropagation}.
This is true; however, backpropagation is not the same thing as SGD. Backpropagation computes the synapse gradients, while SGD utilizes the gradients to coordinate the synapse updates.

As a result, many researchers have attempted to discover a technique which can compute or approximate the synapse gradients in a biological neural network while respecting the constraints of physics and biology; see e.g. \cite{lee2015difference, scellier2017equilibrium, nokland2016direct}. None of the current approaches, however, can be considered a complete theory of gradient descent learning in biological neural networks.

\subsection{Equilibrium Propagation}

One promising solution to the credit assignment problem is called equilibrium propagation \cite{scellier2017equilibrium}. In equilibrium propagation, one begins with a network which naturally optimizes a function called an energy function. In a rough sense, the network “seeks” low values of the energy in its activations. The process of minimizing this energy function corresponds to inference in the neural network. 

More specifically, the network minimizes the energy function by following the negative gradient of the energy with respect to the neurons’ activations, given some inputs to the network. The prototype of an energy-based model is known as a Hopfield network, which is inspired by the Ising model from statistical physics \cite{hopfield1982neural}.

In equilibrium propagation, the energy function is a modified Hopfield energy, defined in terms of weighted interactions between pairs of neurons. The input to neuron $i$ is denoted $u_{i}$, and its output is defined by a nonlinear function $\rho(u_{i})$ representing the firing rate of that neuron. The weights of the interactions $w_{ij}$ are interpreted as the synapses of the network.

\begin{equation}
    E \equiv \frac{1}{2}\sum_{i}u_{i}^2
    -\frac{1}{2}\sum_{i\neq{}j}w_{ij}\rho(u_{i})\rho(u_{j}) - \sum_{i}b_{i}\rho(u_{i})
\end{equation}

The types of network studied in equilibrium propagation are those which have symmetric (undirected) weights, and \textit{converge} to a static fixed-point of the energy. As a result, these networks can only classify static input patterns, and do not describe biological neural networks whose inputs and network states can have non-fixed-point dynamics, and whose synapses can be both directed and recurrent (i.e. possessing directed cycles).

The energy function, which is a global function of the network, is linked to a cost function of the training dataset via another function called the free energy. The free energy $F$ of the network includes the energy $E$ as well as a term which is proportional to the single-target loss function of the network, denoted $C$, with the goal of training the network such that low-energy configurations of the network correspond to low values of the loss function. If low-energy configurations also correspond to low values of the loss, then the network will seek configurations which minimize the loss.

\begin{equation}
    F \equiv E + \beta C
\end{equation}

As is standard in deep learning, the network performs learning via stochastic gradient descent on the weights or synapses of the network with respect to the loss function. What makes equilibrium propagation unique and appealing is that unlike in backpropagation, the gradients are estimated in a purely local fashion. This means the gradient information is locally available to each synapse, i.e. it can be expressed entirely in terms of the presynaptic and postsynaptic neuron activities.

The synapse gradients are extracted through a process where the outputs are perturbed in such a way as to perturb the original “free” fixed point of the network. The difference between the “free” and “perturbed” fixed point is used to compute the gradient of the parameters in the network with respect to the loss function, and this is a purely local computation. Here, the superscript $\beta$ refers to the values of the activations during the perturbed phase, and the superscript 0 refers to the values during the free phase (i.e. $\beta = 0$). 

\begin{equation}
    \Delta w_{ij} = \frac{1}{\beta}\left(\rho (u_{i}^{\beta})\rho (u_{j}^{\beta}) - \rho (u_{i}^{0}) \rho (u_{j}^{0})\right)
\end{equation}

Because of the fixed-point requirement of the energy function, equilibrium propagation does not generalize to time-varying inputs and dynamic networks with memory.

In this paper, using techniques from a branch of physics called Lagrangian mechanics, we will describe how equilibrium propagation can be extended to a \textit{much broader} class  of networks with time-varying dynamics and asymmetric (directed) connections. 

This class of networks is not sufficiently general to include the Hodgkin-Huxley neuron circuit, however we believe this work is a major step in that direction. We will see that in order to cover the Hodgkin-Huxley neuron circuit, we must generalize our notion of fractional memristors to the more general case of \textit{generic memristors}.

\subsection{Lagrangian Mechanics}

Optimization is a central concept in both machine learning and physics. In machine learning, it describes how a deep neural network can be trained to optimize some objective function, such as a desired reward or a classification error. In physics, it arises naturally in the context of Lagrangian mechanics, which is a particularly elegant formulation of the natural laws of mechanics \cite{taylor2005classical}.

Specifically, Lagrangian mechanics states that any physical system \textit{extremizes} (i.e. maximizes or minimizes)\footnote{Technically, extremization sets the first variation of the functional to zero, rather than being explicitly maximized or minimized. However, this is precisely the condition we require in our extension to equilibrium propagation: that the first variation of the action be equal to zero.} a quantity called the \textit{action}. This principle is colloqually known as the \textbf{Principle of Least Action}. The principle of least action is a concept which frames the dynamics of a system as optimizing, or "extremizing" a particular functional of the system's coordinates and velocities. For instance, the path taken by light rays can be interpreted as the trajectory with the shortest travel time. Setting the first variation of the action functional equal to zero yields the differential equations governing the system's time evolution: the celebrated Euler-Lagrange equations \cite{taylor2005classical}. 

This idea of packaging the dynamics of an entire system into a single functional, whose variation returns the dynamics, is known as a variational principle. Given its elegance, it is unsurprising that: 

\begin{quote}
    "Today most physicists would be not only willing to accept as axiomatic the existence of a variational principle, but would be also loath to accept any dynamical equations that were not derivable from such a principle" \cite{relativity1967}.
\end{quote}

The idea that Lagrangian mechanics may be used to extract the gradients in biological neuronal networks in a manner analogous to equilibrium propagation was first proposed by \cite{dold2019lagrangian}. However, the general problem of constructing a Lagrangian for a network of asymmetrically connected neurons with dissipative dynamics was not addressed.

In this work, we describe a set of techniques for explicitly constructing Lagrangians which are capable of describing nonlinear, deep dynamical neural networks, as well as lay the theoretical foundation for an extension to biological neuronal networks.

Specifically, using electrical circuit theory and the fractional calculus of variations, we suggest a form for the Lagrangian of the network which allows us to describe a rich class of \textbf{dissipative} networks with arbitrary combinations of nonlinear capacitors and inductors, linear resistors and sources, and nonlinear fractional memristors. 

Finally, we rigorously derive a gradient estimator for the synapses in the network using a theorem analogous to the central theorem of Equilibrium Propagation.

\section{Lagrangian Extension to Equilibrium Propagation}

The \textit{action} of a physical system is a functional on the possible paths that the coordinates and velocities of the system could take. In other words, the action takes in some possible trajectory that the system could take and returns a real number. This trajectory doesn't necessarily need to be the one physically taken by the system. In fact, the actual trajectory taken by the system corresponds to the one which extremizes (e.g. minimizes) the action. This is the principle of least action.

The action is generally written as the time integral of a very special function, called a Lagrangian function, of the positions and velocities of certain coordinates describing the dynamics of the system:

\begin{equation}
S = \int_{0}^{t} \mathcal{L}(z(t'), \dot{z}(t'), t') dt'
\end{equation}

Here, $z(t)$ is known as the vector of generalized coordinates of the system and $\dot{z}(t)$ denotes the generalized velocities, which are just the time derivatives of the generalized coordinates.

\textit{Our general approach to compute the gradients} of the network’s synapses will be to link the action functional, which is extremized by the network over all possible trajectories, to the total loss of the network’s inference trajectory to create a time-varying analog of the free energy. Using this analog of the free energy, we will then compute a perturbed trajectory in the direction of the desired response and extract the gradient by comparing the original trajectory to the perturbed trajectory.

\paragraph{Generalized Coordinates.}
The generalized coordinates $z(t)$ and the corresponding generalized velocities $\dot{z}(t)$ are constructed via the approach described in \cite{shragowitz1988set}. In short, the electrical circuit under question is represented as a directed graph, with the edges of the graph representing the electrical components of the network, and the nodes representing the junctions of two or more components. A spanning tree is then selected on the graph, along with a cotree, or a set of corresponding links. 

The fluxes (integrals of voltages) of the tree branches form one part of the generalized coordinates called the cut-set fluxes. The charges (integrals of currents) of the links of the cotree form a complementary part of the generalized coordinates called the loop charges. The generalized velocities then take the form of the cut-set voltages and the loop currents, respectively.

In this way, Kirchhoff's laws, which are a set of linear constraints on the network, can be viewed as Lagrange multipliers on the primitive Lagrangian, which is itself a sum of the individual, unconnected element Lagrangians. Specific circuit topologies (element interconnections) then correspond to a specific instantiation of Kirchhoff's laws. Since the loops and cut-sets of the circuit graph determine Kirchhoff's laws, they define the circuit topology.

In a nutshell, Kirchhoff's laws define the circuit topology, and they alter the optimization of the sum of individual component Lagrangians (i.e. the primitive Lagrangian) by enforcing constraints which are interpreted as Lagrange multipliers.

\section{Dissipative Lagrangian via Fractional Calculus}

Neuronal dynamics are known to be dissipative, and understanding dissipative systems is important to understanding neuronal networks \cite{chua2013memristor}. This presents a challenge to conventional Lagrangian mechanics, because standard Lagrangians are not capable of describing dissipative systems. However, it has recently been shown that dissipation can be incorporated into Lagrangian mechanics through a variety of methods \cite{riewe1996nonconservative, galley2013classical, itoh2011memristor, musielak2009general, bateman1931dissipative}. One of these methods is to use "nonstandard" Lagrangians which include \textit{fractional derivative} terms \cite{riewe1996nonconservative}. This is the method we will study in detail in this work. Other dissipative Lagrangian techniques will be discussed in the section on future work.

Since fractional Lagrangians are  unfamiliar to most researchers, we will spend some time setting up the required machinery to understand them: the \textbf{Fractional Calculus of Variations} \cite{malinowska2012introduction}.

\subsection{Overview of the Fractional Calculus of Variations}

The fractional calculus of variations was originally created for the study of non-conservative systems in physics. In the 1990's, a series of papers initiated by F. Riewe's work \cite{riewe1996nonconservative} circumvented the well-known result of Bauer \cite{bauer1931dissipative} that non-conservative (i.e. dissipative) systems cannot be expressed in terms of a variational principle. 

In fact, implicit in Bauer's theorem\footnote{Other loopholes of Bauer's theorem exist and have led to separate developments of dissipative Lagrangian mechanics.} was the restriction that only integer order time derivatives of the coordinates could appear. Riewe showed that by allowing fractional time derivatives of the coordinates, one could derive the appropriate Euler-Lagrange equations from the variation of a single functional, the (fractional) action. 

While Riewe's original work was not fully rigorous, it has since been strengthened by the formal development of the fractional calculus of variations (FCV). It is important to note that the FCV is a relatively new field, and there are still many open questions. For more details, we point interested readers to the following references: \cite{malinowska2012introduction}, \cite{allison2014variational}.

The most important definitions of the FCV are the left and right Riemann-Liouville (RL) fractional integrals. The left and right RL fractional integrals can be described as a generalization of the Cauchy formula for repeated integration. The left RL fractional integral of order $\alpha$ is given as:

\begin{equation}
    \tensor[_a]{I}{^{\alpha}_{t}}z(t) \equiv \frac{1}{\Gamma(\alpha)} \int_{a}^{t} (t-\tau)^{\alpha - 1} z(\tau)d\tau
\end{equation}

Similarly, the right RL fractional integral of order $\alpha$ is given as:

\begin{equation}
    \tensor[_t]{I}{^{\alpha}_{b}}z(t) \equiv \frac{1}{\Gamma(\alpha)} \int_{t}^{b} (\tau - t)^{\alpha - 1} z(\tau)d\tau
\end{equation}

Here, $\Gamma(\alpha)$ is the typical Gamma function, which extends the factorial. It is apparent from the definition that the left RL fractional derivative is equivalent to a temporal convolution of the \textit{past history} of the coordinate $z(t)$ with a power-law kernel $(t-\tau)^{\alpha-1}$, beginning at initial time $a$. In contrast, the right RL fractional derivative is a temporal convolution of the \textit{future evolution} of the coordinate $z(t)$ with the \textit{time-reversed} kernel, from final time $b$.

To extend these fractional integrals to fractional derivatives, there are two possible approaches, which differ only in the details of their boundary conditions. There are the left and right RL fractional derivatives, and the left and right Caputo fractional derivatives. We will make use of both these derivatives.

The left and right RL fractional derivatives are defined as:

\begin{equation}
    \tensor[_a]{D}{_{t}^{\alpha}}z(t) \equiv \left(\frac{d}{dt}\right)^{n}\tensor[_a]{I}{_{t}^{n-\alpha}}z(t)
\end{equation}

\begin{equation}
    \tensor[_t]{D}{_{b}^{\alpha}}z(t) \equiv \left(-\frac{d}{dt}\right)^{n}\tensor[_t]{I}{_{b}^{n-\alpha}}z(t)
\end{equation}

Here, $n$ is the next highest integer order, $\lceil \alpha \rceil$. So, the main idea to arrive at a derivative of fractional order $\alpha$ is to differentiate an integer number of times, $n$, then perform fractional integration to get the rest of the way to $\alpha$.

The definition of the RL fractional derivative is fairly convenient, however in practice it includes boundary terms of fractional order. To get around these problematic boundary terms, one can introduce the Caputo derivative, which simply exchanges the order of differentiation and fractional integration:

\begin{equation}
    \tensor[^C_a]{D}{_{t}^{\alpha}}z(t) \equiv \tensor[_a]{I}{_{t}^{n-\alpha}}\frac{d^{n}z(t)}{dt^{n}}
\end{equation}

\begin{equation}
    \tensor[^C_t]{D}{_{b}^{\alpha}}z(t) \equiv (-1)^{n} \tensor[_t]{I}{_{b}^{n-\alpha}}\frac{d^{n}z(t)}{dt^{n}}
\end{equation}

By including these fractional derivative operators in the Lagrangian, one can show that dissipative forces can be derived from a variational principle. The action we consider in this work will take the following form \cite{agrawal2002formulation}:

\begin{equation}
    S = \int_{0}^{t} \mathcal{L}(z(t'), \dot{z}(t'), \tensor[^C_a]{D}{_{t'}^{\alpha}}z(t'), t')dt'
\end{equation}

This fractional action can be used to describe nonlinear dissipative systems, including electrical circuits \cite{allison2014variational}. In the following sections, we will show how to construct a Lagrangian which describes a broad class of electrical circuits, and derive a gradient estimator for this class of circuits using the fundamental principles  of equilibrium propagation.

\subsection{Form of the Fractional Lagrangian}

We will now construct a simple fractional Lagrangian using known results from the literature. This Lagrangian will  allow us to describe arbitrarily connected, directed networks containing the following two-terminal components:

\begin{itemize}
    \item Linear and Nonlinear Capacitors
    \item Linear and Nonlinear Inductors
    \item Nonlinear Fractional Memristors (Order 1/2)
    \item Linear Resistors
    \item Voltage and Current Sources
\end{itemize}

We begin by describing what we mean by fractional memristor of order 1/2 \cite{machado2013fractional}. In this framework, we use charge $q(t)$ (time integral of current) and flux $\Phi(t)$ (time integral of voltage) as the circuit coordinates, which allows us to write the velocities as voltage $v(t)$ and current $i(t)$. We recall that a (pure) memristor is an element whose constitutive relationship relates flux to charge, while a (nonlinear) resistor relates the first time derivative of flux (i.e. voltage) to the first time derivative of charge (i.e. current). 

Our 1/2 order fractional memristor will relate the 1/2 time derivative of flux, which we denote $\Psi(t)$ to the 1/2 time derivative of charge, which we denote $r(t)$. In this context, it can be intuitively understood as an element which sits "half-way" between a nonlinear resistor and a memristor.

\begin{equation}
    \Psi(t) = \tensor[^C_a]{D}{_{t}^{1/2}}\Phi(t)
\end{equation}

\begin{equation}
    r(t) = \tensor[^C_a]{D}{_{t}^{1/2}}q(t)
\end{equation}

The above definitions for the fractional velocities $\Psi(t)$ and $r(t)$ now allow us to define the constitutive relationship of a 1/2 order memristor, which has one of the following two forms, depending on whether it is controlled by $\Psi$ or $r$:

\begin{equation}
    r(t) = \hat{r}(\Psi(t))
\end{equation}

\begin{equation}
    \Psi(t) = \hat{\Psi}(r(t))
\end{equation}

Now that we have defined the constitutive equation for our fractional memristor, we can  construct the \textit{primitive} Lagrangian, which is a sum of the individual Lagrangians for each component in the network, i.e. each capacitor, resistor, etc., which describes that component's characteristic response. Let us consider the primitive Lagrangian for a simple circuit containing three elements: a nonlinear inductor, a nonlinear capacitor, and a nonlinear fractional memristor of order 1/2.

We use the electrical engineering convention of denoting the unit imaginary by $j$ to avoid confusion with the current $i$. Complex numbers appear naturally in both fractional calculus and electrical circuit theory, and we will discuss their implications below.

\begin{equation}\label{Lagrangian}
    \mathcal{L} = -\int_{0}^{\Phi_{L}}\hat{i}(\Phi')d\Phi' + \frac{j}{2}\int_{0}^{\Psi_{M}}\hat{r}(\Psi')d\Psi' - \int_{0}^{v_{C}}\hat{q}(v')dv'
\end{equation}

The functions $\hat{i}$, $\hat{r}$, and $\hat{q}$ in the Lagrangian are the \textit{Constitutive Equations} of the individual elements. The variables $L$, $M$, and $C$ denote the inductor, the fractional memristor, and the capacitor, respectively. More elements can be included in the circuit by simply adding their constitutive equations to the primitive Lagrangian in the fashion above.

To describe the interconnection of a directed network using these elements, we will add to the primitive Lagrangian a set of Lagrange multipliers corresponding to Kirchoff's laws (i.e. a set of linear constraints on the currents and voltages).

The procedure of transforming the primitive Lagrangian to the full, interconnected Lagrangian is a fairly involved, but algorithmic process \cite{shragowitz1988set}, so we will not describe it in full detail here. Instead, we provide a brief overview:

In short, one must first determine the directed graph formed by the circuit, and select a spanning tree and cotree. Then, using the tree branches and cotree links, construct a set of generalized coordinates (called cut-set fluxes and loop charges), and finally add the corresponding Lagrange multipliers (Kirchhoff constraints) to the primitive Lagrangian.

It is important to note that every \textit{dissipative} element in this analysis is a nonlinear fractional memristor, since capacitors and inductors are not dissipative. However, these nonlinear nonlinear memristors are also capable of describing linear resistors, since a linear memristor of any order is identical to a linear resistor. Since voltage and current sources are just linear resistors with special constitutive equations (horizontal or vertical I-V curves), they can also be described by the fractional memristor term. These can also be understood as standard linear dissipative elements treated via the fractional variational approach in \cite{allison2014variational, agrawal2002formulation}.

Hence, we have constructed a dissipative Lagrangian capable of describing circuits with nonlinear (and linear) capacitors and inductors, nonlinear fractional memristors, and linear resistors and sources.

We will now restate the theorem available in \cite{agrawal2002formulation} which shows that the variation of the action defined by the fractional Lagrangian of equation \ref{Lagrangian} indeed yields the correct Euler-Lagrange equations describing the dynamics of the circuit.

\subsection{Variation of the Fractional Action}

First, we return to the form of the action $S$. For simplicity, we use a single coordinate $\Phi$ describing a single node in the circuit. This corresponds to the total flux (integrated voltage) at the node. The velocity $\dot{\Phi} = v$ then corresponds to the node voltage, and the 1/2 fractional derivative intuitively represents a variable which sits "half-way" between flux and voltage, and represents the nonlinear dissipation and fading memory effects of a fractional memristor. Multiple nodes can be handled in a similar manner.

\begin{theorem}\label{thm1}
Let S an action of the form

\begin{equation}\label{action}
    S = \int_{a}^{b} \mathcal{L}(\Phi, \dot{\Phi}, \tensor[^C_a]{D}{_{t}^{1/2}}\Phi, t)dt
\end{equation}

Where the function $\Phi(t)$ satisfies the fixed boundary conditions $\Phi(a) = \Phi_{a}, \Phi(b) = \Phi_{b}, \dot{\Phi}(a) = \dot{\Phi}_{a}, \dot{\Phi}(b) = \dot{\Phi}_{b}$, and let $\mathcal{L} \in C^{2}[a,b] \times \mathbb{R}^{4}$. Then, the necessary condition for $S$ to possess an extremum at $\Phi(t)$ is that the function $\Phi(t)$ fulfills the following fractional Euler-Lagrange equation:

\begin{equation}
    \frac{\partial \mathcal{L}}{\partial \Phi} - \frac{d}{dt} \frac{\partial \mathcal{L}}{\partial \dot{\Phi}} + \tensor[_t]{D}{_{b}^{1/2}} \frac{\partial \mathcal{L}}{\partial \left(\tensor[^C_a]{D}{_{t}^{1/2}}\Phi\right)} = 0
\end{equation}
\end{theorem}

The proof of a more general version of Theorem \ref{thm1} can be found in \cite{agrawal2002formulation}. The first two terms can be immediately identified as the "usual" Euler-Lagrange terms describing the generalized force and generalized momentum for coordinate $\Phi$ and velocity $\dot{\Phi} = v$. The third term is associated with a nonlinear dissipative element, which we will identify with the fractional memristor in the context of electrical circuits. 

To further elucidate the relationship between the fractional term in the Euler-Lagrange equation and the fractional memristor, we consider the Lagrangian of equation \ref{Lagrangian}, and compute the third term in the Euler-Lagrange equation:

\begin{equation}\label{memristor}
    \tensor[_t]{D}{_{b}^{1/2}} \frac{\partial \mathcal{L}}{\partial \left(\tensor[^C_a]{D}{_{t}^{1/2}}\Phi \right)} = \tensor[_t]{D}{_{b}^{1/2}} \frac{\partial \mathcal{L}}{\partial \Psi} = \tensor[_t]{D}{_{b}^{1/2}} \left( j \hat{r}(\Psi_{M}) \right) = \tensor[_t]{D}{_{b}^{1/2}} \left(j \tensor[^C_a]{D}{_{t}^{1/2}}q(t)\right)
\end{equation}

We now use two relations, stated below, to simplify the above equation:

\begin{equation}\label{leftright}
    j\tensor[^C_a]{D}{_{t}^{1/2}} x(t) = \tensor[^C_t]{D}{_{b}^{1/2}} x(t) \text{ in the limit } a \rightarrow b
\end{equation}

\begin{equation}\label{semigroup}
    \tensor[_t]{D}{_{b}^{1/2}} \left(\tensor[^C_t]{D}{_{b}^{1/2}} x(t)\right) = \frac{d}{dt} x(t)
\end{equation}

The proof of equation \ref{leftright} can be found in \cite{klimek2001fractional} and equation \ref{semigroup} follows from the general semigroup property of fractional integrals \cite{diethelm2010analysis}.

Returning to our derivation of the fractional memristor term in the Euler-Lagrange equation, we can simplify equation \ref{memristor} as:

\begin{equation}
    \tensor[_t]{D}{_{b}^{1/2}} \left( j \tensor[^C_a]{D}{_{t}^{1/2}}q(t) \right) = \frac{d}{dt} q(t) = i(t)
\end{equation}

This allows us to express Kirchoff's current law for the node(s) of the circuit given the nonlinear constitutive equation(s) relating $\Psi(t)$ and $r(t)$. The other constitutive elements in the circuit (i.e. capacitors, inductors) are then related to each other via their contributions to the total node current, which sums to zero. This yields the final differential equation(s) of the circuit.

We have therefore established that a fractional action and corresponding fractional Lagrangian can be defined for circuits containing nonlinear 1/2 order fractional memristors (and therefore linear resistors and sources), nonlinear capacitors, and nonlinear inductors. Now, we derive a gradient estimator for these circuits which can be used to perform stochastic gradient descent (SGD) on the parameters of the network (i.e. linear resistors / synapses).

\section{Derivation of the Gradient}

We will now separate the Lagrangian into sets of terms which will aid in our analysis. We assume the network takes the form of a deep neural network, similar to our previous work \cite{kendall2020training}, with layers of nonlinear, \textit{dynamic} subcircuits (neurons) separated by fully connected layers of linear resistors (synapses). 

\subsection{Structure of the Network}

We will represent the outputs of the network as linear capacitors, with the voltages on the upper plates of the capacitors giving the neural network's outputs $v_{k}(t)$. The bottom plate voltages of the output capacitors will be given by $T_{k}$ which represent the targets. The inputs will be given by independent, time-varying voltage sources $V_i$. The conductances of the synapses will be denoted by $g_l$, and are assumed to be constant between updates. 

All other elements of the network, including the nonlinear fractional memristors and capacitors of the hidden neurons, we lump into a single term $\mathcal{L}_H$ for simplicity. We do this, because the Lagrangian is sum separable in each of its circuit elements, and we only require the gradients of the linear synapses. Note that $\mathcal{L}_H$ also takes the form given by the Lagrangian in equation \ref{Lagrangian}.

We also remark that the linearity of the synapses and capacitors allows them to be written as a sum of squares of the constitutive equations, rather than a sum of integrals. Therefore, we can write the total Lagrangian of the system as:

\begin{equation}
    \mathcal{L} = \sum_{l}^{L} \frac{j}{2} g_{l}\left(\tensor[^C_a]{D}{_{t}^{1/2}} \Phi_l\right)^2 - \sum_{k}^{K} \beta C(v_{k}(t) - T_{k}(t))^2 + \mathcal{L}_{H}
\end{equation}

Where $\beta C(v_{k}(t) - T_{k}(t))$ are the constitutive equations of the $K$ \textit{linear} output capacitors, each with capacitance $\beta C$.

The Lagrangian is now linked to the network loss $J$ by considering the voltage drops across the $K$ output capacitors. If $J$ is given by the mean squared error (MSE) loss, then we have that:

\begin{equation}
    \frac{\partial S}{\partial \beta} = -C \int_{a}^{b} \sum_{k} (v_{k}(t) - T_{k}(t))^2 dt = -CJ
\end{equation}

Which is the MSE loss times a constant factor $-C$. We can also extract the partial derivatives of the action with respect to the (constant) resistances $R_l$:

\begin{equation}
    \frac{\partial S}{\partial g_l} = \frac{\partial}{\partial g_l} \int_{a}^{b} \frac{j}{2} g_{l} \left(\tensor[^C_a]{D}{_{t}^{1/2}}\Phi_{l} \right)^2 = \int_{a}^{b} \frac{j}{2} \left( \tensor[^C_a]{D}{_{t}^{1/2}}\Phi_{l} \right)^2
\end{equation}

Now that we have the above partial derivatives, we will use the fundamental principles of equilibrium propagation to derive an equation for the gradient.

\subsection{Form of the Gradient}

In order to derive the gradient, we will exploit the relationship between the action functional $S$ over trajectories and the loss functional $J$ over trajectories.

The goal is to compute the gradient of the loss functional of a trajectory with respect to each parameter (synapse) $g_l$ in the network.

We first begin with the extremization condition of the action, which generalizes the fixed-point condition of equilibrium propagation. The extremization condition states that the first variation of the action along any trajectory \textit{actually taken} by the network is always zero:

\begin{equation}
    \delta S = \int_{a}^{b} \frac{\delta S}{\delta \Phi(t)} \delta \Phi(t) dt = 0
\end{equation}

For any admissible\footnote{i.e. does not violate Kirchoff's laws or the constitutive equations} variation of the coordinates $\delta \Phi$. This can be re-written in the following form, for arbitrary $\delta \Phi(t)$:

\begin{equation}
    \frac{\delta S}{\delta \Phi(t)} = 0
\end{equation}

It is this condition, the vanishing of the first variation with respect to the dynamical variables, that results in the Euler-Lagrange equations. Here, we show that this fact can be exploited to derive the gradient.

Note that a small change in $\beta$ or $g_l$ will result in a first-order variation of the coordinate trajectory of the network $\delta \Phi(t)$. This corresponds to the "indirect" influence of $g_{l}$ or $\beta$ on the action, via the system dynamics. We can use this observation with the extremization condition of the action to calculate the total derivative of $S$ with respect to $\beta$ and $g_l$. 

\begin{equation}
    \frac{dS}{d\beta} = \frac{\partial S}{\partial \beta} + \frac{\delta S}{\delta \Phi(t)} \frac{d(\delta \Phi(t))}{d\beta} = \frac{\partial S}{\partial \beta}
\end{equation}

Because the variation of the action with respect to first-order variations in the coordinate trajectory is equal to zero by the extremization condition, we see that the total derivative of the action with respect to $\beta$ is equal to the partial derivative. Similarly, for $g_l$:

\begin{equation}
    \frac{dS}{d g_l} = \frac{\partial S}{\partial g_l} + \frac{\delta S}{\delta \Phi(t)} \frac{d(\delta \Phi(t))}{d g_l} = \frac{\partial S}{\partial g_l}
\end{equation}

The extremization condition leads to the RHS of the above equations for any $\delta \Phi(t)$ induced by a small change in $\beta$ or $g_l$. Therefore:

\begin{equation}
    \frac{d}{dg_l} \frac{\partial S}{\partial \beta} = \frac{\partial^{2} S}{\partial g_l \partial \beta} = \frac{\partial^{2} S}{\partial \beta \partial g_l} = \frac{d}{d\beta} \frac{\partial S}{\partial g_l}
\end{equation}

We now substitute the formulas derived above for the partial derivatives of the action with respect to $\beta$ and $R_l$ to get:

\begin{equation}
    \frac{d}{dg_l} \frac{\partial S}{\partial \beta} = -C \frac{dJ}{dg_l} = \frac{d}{d\beta} \frac{j}{2} \int_{a}^{b} \left(\tensor[^C_a]{D}{_{t}^{1/2}} \Phi_l\right)^2 dt
\end{equation}

Rearranging, we arrive at the final form of the gradient:

\begin{equation}\label{gradient}
    \frac{dJ}{dg_l} = \frac{-j}{2C} \frac{d}{d\beta} \int_{a}^{b} \left(\tensor[^C_a]{D}{_{t}^{1/2}} \Phi_l\right)^2 dt
\end{equation}

The above equation can be approximated by computing two trajectories of the network: a first, unperturbed trajectory (with $\beta = 0$, and a second trajectory where the outputs are perturbed in the direction of the desired targets $\beta = \beta$). The difference between these two trajectories can then be fractionally integrated, as in equation \ref{gradient}, to arrive at the final form of the update rule:

\begin{equation}\label{update}
    \Delta g_l = \frac{j}{2C} \lim_{\beta \rightarrow 0} \frac{1}{\beta} \left[ \int_{a}^{b} \left(\tensor[^C_a]{D}{_{t}^{1/2}} \Phi_l(\beta, t) \right)^2 dt - \int_{a}^{b} \left(\tensor[^C_a]{D}{_{t}^{1/2}} \Phi_l(0, t) \right)^2 dt \right]
\end{equation}

The form of the above learning rule is that of \textit{trajectory learning}. We do not comment in detail here on how such trajectories can be evaluated in the context of real neural circuits, however we can imagine several possibilities:

\begin{itemize}
    \item One possibility is that during a \textit{wake} phase, samples are taken from the environment which correspond to the unperturbed trajectories, and the fractional integrals are carried out, making small increments to the weights. Then, during a \textit{sleep} phase, trajectories are drawn from a generative model to serve as contrastive examples, similar to a Boltzmann machine \cite{hinton2007boltzmann}.
    
    \item Another possibility is that short snippets of experience are stored in a replay memory, analogous to the hippocampus, and played back during sleep to give the perturbed trajectories. This is similar to methods used in deep reinforcement learning \cite{mnih2013playing}
    
    \item Still another method is that value estimates are predicted along each trajectory, corresponding to error neurons. The difference between the actual activation of a neuron and the predicted value would then drive learning in a fully online fashion, similar to predictive coding techniques \cite{spratling2017review}
\end{itemize}

We also note that the formula for the update contains the imaginary unit $j$. Two possibilities for handling the imaginary unit are given. The first is that the output capacitors be replaced with output resistors. In this way, the two terms containing $j$ in the gradient formula will multiply together, and the imaginary unit will cancel out. 

An issue with this approach is that it complicates the formula for the update rule, because one needs to perform a fractional integration in order to evaluate the loss. However, it is possible to do this in practice.

A second interpretation is that the imaginary unit should be understood in the context of the Laplace Transform of the circuit. It is well-known that fractional derivatives are most naturally computed in the Laplace domain, where fractional differentiation of order 1/2 corresponds to multiplication in the complex frequency domain by $s^{1/2}$, and fractional integration of order 1/2 corresponds to multiplication in the complex frequency domain by $s^{-1/2}$. This follows from the fact that fractional differentiation and integration are convolution operators.

The interpretation here is that the imaginary part of the weight update represents a phase shift in complex frequency domain of the network's dynamics, or a time delay in the time domain.

\subsection{Extension to Hodgkin-Huxley Circuit}

It has been shown that circuits with nonlinear capacitors, inductors, and fractional memristors (including linear resistors and sources) can be modeled using a variational principle via the fractional calculus of variations.

In \cite{chua2012hodgkin}, Chua showed that the "time-varying conductances" of the Hodgkin-Huxley neuron circuit are actually memristors. This suggests, since networks of Hodgkin-Huxley neurons contain only linear capacitors, resistors, and sources, along with (first order) nonlinear memristors, that an appropriate extension of the above framework should be capable of describing these neural circuits.

As shown in Chua's work, the sodium and potassium channel memristors of the Hodgkin-Huxley neuron are actually \textit{generalized} memristors, in that they cannot be described as a constitutive relation between charge and flux. These generalized memristors can be described by one or more internal state variables, and these internal state variables are outside of the scope of the analysis presented here.

If generalized memristors are able to be described by fractional Lagrangian mechanics, then we could derive the gradients of such a circuit using the techniques presented above. Therefore, we consider this paper to be a \textit{call to action} to construct a Lagrangian which is capable of generating the equations of motion of more general memristive systems.

\section{Conclusion and Future Work}

In this work, we have derived a gradient estimator for a broad class of circuits including nonlinear capacitors, inductors, and (1/2 order) fractional memristors, as well as linear resistors and independent (voltage and current) sources. We have used the methods of equilibrium propagation combined with fractional Lagrangian mechanics to derive this gradient estimator. To our knowledge, this is the first time the methods of equilibrium propagation have been extended to a class of asymmetrically connected networks with non-fixed-point dynamics. 

While we have considered in this work only the fractional calculus approach to extending Lagrangian mechanics to dissipative nonlinear systems, it is by no means the only technique to do this. There are several orthogonal research directions which aim to explicitly incorporate nonlinear dissipative dynamics within the Lagrangian formalism.

Perhaps the most rigorous of these approaches is by considering that all physical systems are in fact conservative if one includes the microscopic variables of the system. For instance, by including a collection of harmonic oscillators, one can model dissipative effects \cite{caldeira1983path}. The process of integrating out these microscopic variables is highly complex, however, and arbitrary nonlinear dissipative behavior is difficult to model in such a system.

Additionally, one can take the approach of \cite{itoh2011memristor}, where a change of variables is employed to allow the explicit inclusion of dissipative memristive elements. This approach creates a "pseudo-conservative" system with reversible temporal dynamics, which can be described using a principle of least action.

Finally, there is the technique of doubling the number of degrees of freedom of the system. One set of DoF corresponds to the forward time evolution of the system, while the mirror set of DoF corresponds to unphysical variables which are ignored in the analysis, and track the transfer of energy to the environment \cite{bateman1931dissipative}.

We hope this work will serve as inspiration for a more general Lagrangian, which is capable of describing a broader range of circuits, especially including the Hodgkin-Huxley neuron ciruit or circuits with multi-terminal elements such as transistors.

\section*{Acknowledgments}

The author would like to thank Yoshua Bengio, Benjamin Scellier, Ross Pantone, and Juan Nino for their helpful insights and feedback on the work contained in this manuscript.

\bibliographystyle{ieeetr}  
\bibliography{biblio}

\end{document}